\documentclass{article}

\PassOptionsToPackage{numbers, sort}{natbib}

    \usepackage[final]{neurips_2024}

\usepackage[utf8]{inputenc} %
\usepackage[T1]{fontenc}    %
\usepackage{hyperref}       %
\usepackage{url}            %
\usepackage{booktabs}       %
\usepackage{amsfonts}       %
\usepackage{nicefrac}       %
\usepackage{microtype}      %
\usepackage{xcolor}         %

\usepackage{amsmath}
\usepackage{amssymb}
\usepackage{mathtools}
\usepackage{amsthm}
\usepackage{wrapfig}

\usepackage{graphicx}
\usepackage{subfigure}

\usepackage{enumitem}

\newenvironment{denseenum}{
\begin{enumerate}[topsep=2.5pt, partopsep=0pt, leftmargin=1.5em]
  \setlength{\itemsep}{2.5pt}
  \setlength{\parskip}{0pt}
  \setlength{\parsep}{0pt}
}{\end{enumerate}}

\title{Learn To be Efficient: Build Structured Sparsity in Large Language Models}

\author{%
  Haizhong Zheng\,\,\footnotemark[2] \And Xiaoyan Bai\footnotemark[2] \And Xueshen Liu\footnotemark[2]\\
  \AND Z. Morley Mao\footnotemark[2] \And Beidi Chen\footnotemark[3] \And Fan Lai\footnotemark[4]
  \And Atul Prakash \footnotemark[2]
  \\
  \AND
  \textnormal{\footnotemark[2]\,\,University of Michigan
  \footnotemark[3]\,\;Carnegie Mellon University} \\
  \textnormal{\footnotemark[4]\,\; University of Illinois Urbana-Champaign} \\
  \texttt{\{hzzheng, smallyan, liuxs, zmao, aprakash\}@umich.edu ,} \\
  \texttt{beidic@andrew.cmu.edu, fanlai@illinois.edu}
}

\begin{document}

\maketitle

\begin{abstract}

Large Language Models (LLMs) have achieved remarkable success with their billion-level parameters, yet they incur high inference overheads. The emergence of activation sparsity in LLMs provides a natural approach to reduce this cost by involving only parts of the parameters for inference. However, existing methods only focus on utilizing this naturally formed activation sparsity in a post-training setting, overlooking the potential for further amplifying this inherent sparsity. In this paper, we hypothesize that LLMs can \emph{learn to be efficient} by achieving more structured activation sparsity. To achieve this, we introduce a novel training algorithm, Learn-To-be-Efficient (LTE), designed to train efficiency-aware LLMs to learn to activate fewer neurons and achieve a better trade-off between sparsity and performance. Furthermore, unlike SOTA MoEfication methods, which mainly focus on ReLU-based models, LTE can also be applied to LLMs like LLaMA using non-ReLU activations. Extensive evaluation on language understanding, language generation, and instruction tuning tasks show that LTE consistently outperforms SOTA baselines. Along with our hardware-aware custom kernel implementation, LTE reduces LLaMA2-7B inference latency by 25\% at 50\% sparsity.
We make our code publicly available at 
\href{https://github.com/haizhongzheng/LTE}{GitHub}\footnote{\url{https://github.com/haizhongzheng/LTE}}.

\end{abstract}

\section{Introduction}

\begin{wrapfigure}{r}{0.45\textwidth}
    \centering
    \includegraphics[width=\linewidth]{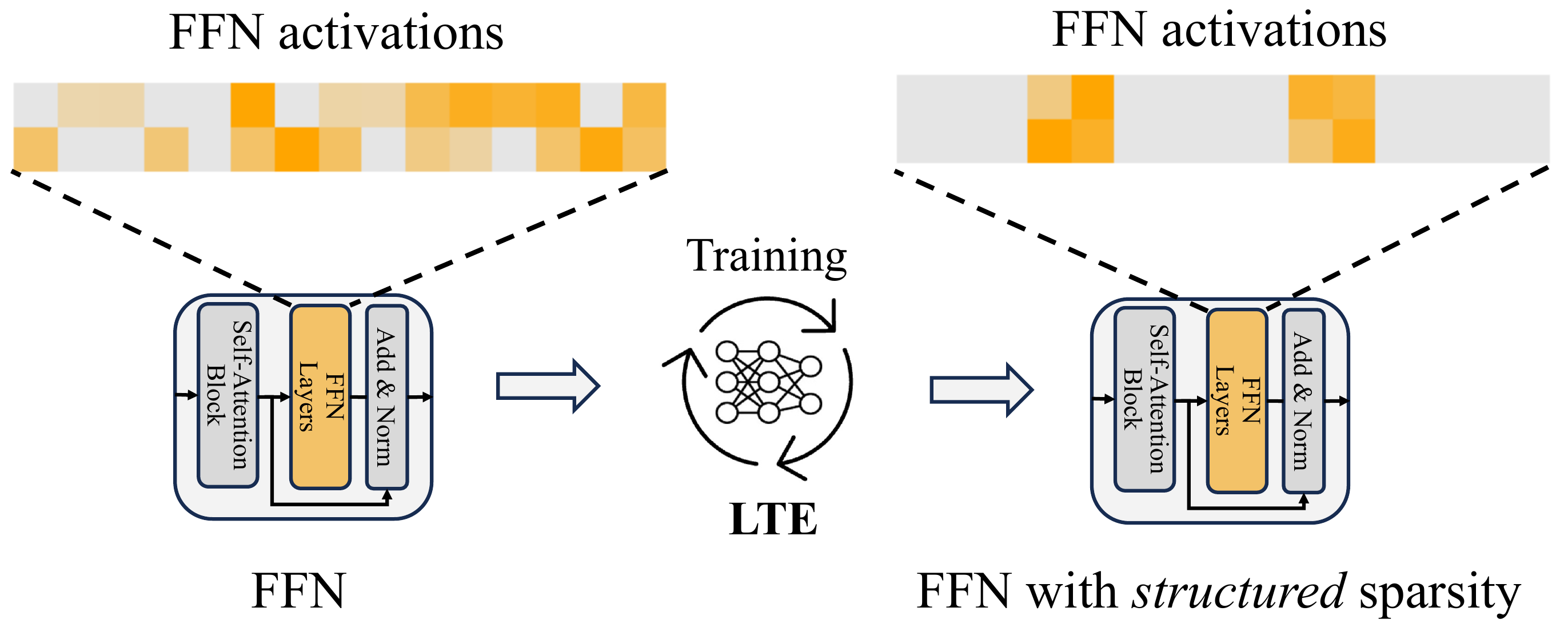}    
    \caption{Learn-To-be-Efficient~(LTE) performs efficiency-aware training to construct structured model contextual sparsity for fast inference. Activated neurons are in orange. }
    \label{fig:lte-idea}
\end{wrapfigure}

The exponential growth in data volumes and model sizes has catalyzed significant breakthroughs in large-scale models, enabling a wide range of applications~\cite{brown2020language, zhang2022opt, zheng2017smoke, zhang2014you, redmon2016you}.
Among them, large language models (LLMs), like GPT-3~\cite{brown2020language}, OPT~\cite{zhang2022opt}, and LLaMA~\cite{touvron2023llama, touvron2023llama2}, have demonstrated impressive natural language ability. 
However, the skyrocketing number of model parameters~\cite{liu2023deja, sun2023simple} and dataset size~\cite{zheng2022coverage, xia2024less, zheng2023leveraging, wu2024self}
has introduced challenges in further scaling those models.
The exponential growth in model size has not only inflated the deployment costs of LLMs, due to their significant computational and memory demands during inference, but also affected the user experience in many latency-sensitive applications, such as chatbots~\cite{achiam2023gpt} and autonomous driving~\cite{li2023towards, sun2023calico}. 
Recent advances have been leveraging sparsity to improve LLM inference efficiency, including model weight quantization~\cite{frantar2022gptq} and pruning~\cite{ma2023llm}, token sparsity~\cite{zhang2023h, xiao2023efficient}, and activation sparsity~\cite{liu2023deja, zhang2022moefication}.

During LLM inference, Feed Forward Network (FFN) layers are the primary efficiency bottleneck, accounting for over 60\% of the FLOPs and I/O operations~\cite{liu2023deja}, but they exhibit high activation sparsity, especially in ReLU-based LLMs~\cite{li2023lazy}.
For example, in a 175B OPT model, more than 95\% of activations in FFN layers are zeros~\cite{liu2023deja}. 
Recent advances leverage this activation sparsity via \emph{MoEfication}~\cite{zhang2022moefication, mirzadeh2023relu}: they convert FFN layers to MoE layers by grouping neurons in the FFN intermediate layer into $n$ experts and then applying a router to select $k$ most important experts for each input token. Unlike training the MoE model from scratch, MoEfication converts FFN layers in pretrained LLMs into MoE layers with relatively small training resources.

However, existing advances only focus on manipulating the pre-trained activation sparsity of neurons, overlooking the potential for further increasing this inherent sparsity. 
Moreover, they are limited to the historically ReLU-based LLMs, whereas emerging advanced models use soft activation functions for better model quality (e.g., SwiGLU in LLaMA~\cite{shazeer2020glu} and GeGLU in Gemma~\cite{team2024gemma}), exhibiting much lower natural activation sparsity. 
While existing works suggest replacing the model activation with ReLU to enable sparsity~\cite{zhang2022moefication, mirzadeh2023relu}, we show that this can hurt model performance~(Section~\ref{ssec:performance-comparison}).

In this paper, we hypothesize that LLMs can learn to be efficient and achieve more structured activation sparsity.
As shown in Figure~\ref{fig:lte-idea}, our key insight is that, without compromising model quality, \emph{we can also train LLMs to be efficiency-aware by developing more structured sparsity, which is more friendly for achieving hardware speedup.}
However, creating structured sparsity in pretrained LLMs is non-trivial due to two practical challenges:  

\begin{denseenum}

\item 
\emph{How to train routers more stably?}
The widely-used Top-k Softmax routing can lead to a severe accuracy drop~(Section\ref{ssec:noisy-top-k}). How to jointly train the model and routers in a MoEfication setting is still open-ended.

\item 
\emph{How to select the right experts in serving?}
The number of experts needed in MoE layers depends on specific inputs and layers, implying a trade-off between model efficiency and quality.
\end{denseenum}

To address these challenges, we introduce Learn-To-be-Efficient (LTE), a novel training algorithm to train efficiency-aware models. 
LTE integrates an efficiency loss penalty, encouraging models to activate fewer neurons in their FFN layers while keeping good task performance. 
Additionally, LTE adopts a threshold-based Sigmoid routing strategy to select experts and employs a two-stage training mechanism to improve training stability.
LTE achieves a more flexible selection of experts instead of selecting a fixed number of experts for all layers and inputs.
Same as Deja Vu~\cite{liu2023deja} and moefication~\cite{zhang2022moefication}, LTE provides very structured sparsity. 
We further develop a custom CUDA kernel with Triton, a Python-like CUDA programming language, to enable wall-clock time speedup from such structured sparsity~(Section~\ref{ssec:hardware-implementation}).

We evaluate LTE on both encoder-based models~(RoBERTa$_\text{base}$, RoBERTa$_\text{large}$~\cite{liu2019roberta}) 
and decoder-based models~(GPT2-Medium~\cite{radford2019language}, LLaMA2-7B~\cite{touvron2023llama}) with the HuggingFace's transformers.
Our extensive experiments on Natural Language Understanding~(NLU) tasks, downstream Natural Language Generation~(NLG) tasks, and instruction tuning tasks,  show that LTE consistently outperforms state-of-the-art designs. 
For instance, LLaMA with LTE provides a 1.83x - 2.59x FLOPs speed-up on NLG tasks without compromising model quality.
After integrating our hardware-efficient implementation of sparse matrix multiplication kernel,
LTE reduces LLaMA2-7B 25\% wall-clock time latency at around 50\% sparsity.
Our evaluation results demonstrate that LTE effectively builds more structured sparsity even on LLMs with soft activation functions.

\section{Related Work}
\label{sec:lte-related-work}
\textbf{Mixture of Experts (MoE).}
MoE was proposed by~\cite{6797059} a few decades ago to build an integral system with subset networks. 
Recently, MoE layers have been used in the Transformer architecture as a substitute for the MLP block~\cite{shazeer2018mesh, fedus2022switch}, with the recent Mistral-7B model~\cite{jiang2023mistral} as a prominent example.
In MLP blocks of MoE transformers, instead of using an FFN layer for calculating output, MoE layers construct multiple smaller FFN layers and employ a router to choose a subset of experts to do conditional computation.
Even though transformers can have billions of parameters, only a subset of experts are activated for each token, thus effectively increasing model activation sparsity (i.e., the number of skipped neurons in execution).

\textbf{LLM Contextual Sparsity.} 
Recent studies~\cite{li2023lazy, liu2023deja} show that trained ReLU-based transformers naturally show a great sparsity in the activation of FFN layers. 
For example, in a 175B OPT model, more than 95\% activations in FFN layers are zeros~\cite{liu2023deja}. While SOTA models are increasingly using diverse soft activations, like GeLU and SwiGLU, existing work shows that replacing soft activations with ReLU and fine-tuning the model can increase the activation sparsity without greatly hurting model quality~\cite{zhang2022moefication, mirzadeh2023relu, li2023lazy}.
Yet, we find that this benefit is not consistent for all datasets (Section~\ref{ssec:performance-comparison}).
This emerging sparsity indicates that a large portion of neurons are unnecessary in LLM inference (i.e., sparsity), which we can leverage to improve LLM inference efficiency like using MoEfication~\cite{zhang2022moefication}.
Recent papers~\cite{liu2023deja, zhang2022moefication} show that a single FFN layer in pretrained LLMs can be converted to an MoE layer, by splitting the matrices of FFN layers into exclusive sub-matrices and employing a router to activate only neurons in a subset of experts.
Similar to MoE models, MoEfication can effectively accelerate LLM inference by only using part of the parameters~\cite{zhang2022moefication, liu2023deja}.

\textbf{Model Weight Sparsity.} 
Another orthogonal direction regarding model sparsity is static model weight sparsity~\cite{frantar2023sparsegpt, sun2023simple}, which often prunes the model weights and keeps them static during inference. 
For instance, Wanda~\cite{sun2023simple} estimates model weight importance and prunes unimportant weights, so all inputs will use the same subset of weights. In contrast, contextual sparse models select a different subset of weights for different inputs.
Nonetheless, model pruning can be applied to contextual sparse models too, meaning a complementary optimization to contextual sparsity.

\vspace{-0.2cm}
\section{Background and Motivation}
\vspace{-0.1cm}
In this section, we first briefly introduce the MoEfication of FFN layers in transformers. 
Then, we present a study on the limitation of applying noisy top-K Softmax routing in the Moefication setting.

\subsection{Background: MoEfication}

MoEfication~\cite{zhang2022moefication} is a way to group neurons in FFN layers, thereby converting FFN layers to MoE layers for better execution speedup. 
For a transformer whose hidden states and FFN intermediate dimension are $d_{model}$ and $d_{FFN}$, respectively, the FFN layers process the input as follows:
\[h = xW_{1} + b_2\]
\[FFN(x) = \sigma(h)W_{2} + b_2\]
where $x \in \mathbb{R}^{d_{model}}$, $W_1, W_2$ are weight metrics, $b_1, b_2$ are biased term and $\sigma$ is an activation function.
MoEfication aims to group $d_{FFN}$ neurons in the FFN intermediate layer into $n$ experts.
Then, a router is trained for each FFN layer to select $k$ experts (k < n), thus reducing activation load, to speed up inference.
A recent work, Deja Vu~\cite{liu2023deja}, uses a similar setting but treats each neuron as an expert.

However, SOTA MoEfication methods overlook the potential for further optimizing the activation sparsity of LLMs. 
Besides, for LLMs employing soft activations such as GeLU~\cite{hendrycks2016gaussian} and SwiGLU~\cite{shazeer2020glu}, 
MoEfication~\cite{zhang2022moefication} proposes to replace soft activations with ReLU and fine-tune the model to improve activation sparsity. 
Although recent works~\cite{zhang2022moefication, mirzadeh2023relu} show that this replacement has a marginal impact on model performance, we notice that this hypothesis is not widely applicable~(Section~\ref{ssec:performance-comparison}).

\subsection{Limitations of Noisy Top-K Softmax Routing} 
\label{ssec:noisy-top-k}

\begin{wrapfigure}{r}{0.5\textwidth}
    \centering
    \includegraphics[width=\linewidth]{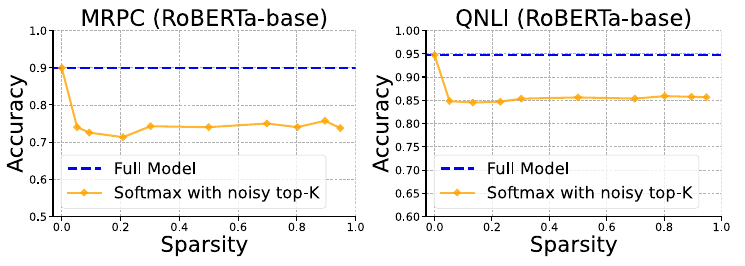}    
    \caption{
    Models trained with noisy top-K Softmax routing experience severe accuracy drops, even at very low levels of sparsity. }
    \label{fig:ntopk}
\end{wrapfigure}

One potential solution to train and convert the model into MoE layers is noisy top-K Softmax routing~\cite{fedus2022switch, shazeer2016outrageously}.
It selects $k$ experts based on the highest router outputs and then calculates Softmax values for these selected outputs. 
The output of the MoE layer is determined by summing the products of these softmax values with their corresponding experts' outputs.
Additionally, a small Gaussian noise is added to the router's outputs during training to encourage exploration across different experts, which addresses the issue of unselected experts being non-differentiable.

To study the effectiveness of noisy top-K Softmax routing, we follow the standard MoEfication setting~\cite{zhang2022moefication} to form experts in FFN layers: we fine-tune a model on a downstream dataset and group neurons into experts via parameter clustering.
Subsequently, we integrate Softmax routers into models and jointly train both the model and routers.
As shown in Figure~\ref{fig:ntopk}, we observe noticeable accuracy drops even at very low sparsity levels.
When diving into the expert scores produced by routers, we observe a very biased expert score allocation: only one expert in an MoE layer has an expert score close to 1, while other experts receive nearly 0 scores.
This results in only one expert contributing to the inference in each MoE layer, leading to performance drops. 

A potential reason behind this biased allocation is that the sum of expert scores in Softmax routing is constrained to 1. 
Unlike traditional MoE models, which typically select 1-2 experts~\cite{shazeer2018mesh, fedus2022switch},
MoEfied models need to select a much larger number of experts~\cite{liu2023deja, zhang2022moefication}, but the sum of experts score is still 1; this can complicate the allocation of the Softmax budget and cause a biased allocation.

\vspace{-0.2cm}
\section{Methodology: Learn To be Efficient}
In this section, we present LTE (Learn To be Efficient) to learn more structured MoE in FFN layers for fast inference.
Specifically, we tackle two key challenges toward practical MoEfication for diverse LLMs:
(1) How to train routers more stably? (Section~\ref{ssec:lte}); and
(2) How to select the right experts in serving? (Section~\ref{sssec:training}).

\subsection{Experts Grouping}
\label{ssec:grouping}
The first step of LTE is to group neurons in FFN layers to form experts.
Expert grouping aims to group neurons that are often activated together, thereby improving efficiency by reducing the total number of activated neurons.
Here, constructing too many experts (e.g., each neuron as an expert) can significantly increase the cost, while too few can lead to poor model quality. Moreover, we need to group neurons without a significant performance drop on the pre-trained model.

Following the previous work~\cite{zhang2022moefication}, we group 32 neurons as an expert in FFN layers, and use parameter clustering to group MLP neurons into different experts.
Fundamentally, in FFN layers, each neuron is associated with a column of $W_1$ (which is a $d_{model}$ vector).
Parameter clustering first treats the corresponding vector in $W_1$ as the feature for a neuron, then applies the balanced K-Means algorithm~\cite{malinen2014balanced} to cluster neurons into $n$ clusters.
Each cluster has $32$ neurons by default and will be treated as an individual MoE expert.\footnote{
For FFN layers using SwiGLU (e.g., LLaMA), there are two linear networks to calculate neuron values: $SwiGLU(x) = (Swish(xW) \otimes xV )W_2.$
We use the weight of the gate network, $W$, to cluster neurons.
}
We also conduct an ablation study on other alternative grouping strategies in Section~\ref{ssec:ablation}.

\subsection{Adaptive Expert Routing}
\label{ssec:lte}

\subsubsection{Router Design}
After constructing the experts, we need to decide on the right routing strategy. Similar to existing MoE designs, we use a fully-connected layer as the router network.
The input to the router is the output of the preceding self-attention block, and the router output is the expert score for each expert.

\textbf{Routing function.}
To address the biased expert score issue (Section~\ref{ssec:noisy-top-k}), we employ the Sigmoid function on router outputs to get expert scores. Unlike the Softmax function, the Sigmoid function computes each expert score independently, which circumvents the biased expert score due to the constraint on the sum of outputs in the Softmax function:
\begin{equation}\label{eq:sigmoid-router}
G(x)_i = Sigmoid(x \cdot W_{g,i}) = \frac{1}{1+e^{-x \cdot W_{g,i}}},
\end{equation}
where $x$ is the input to the corresponding FFN layer, $i$ is the index for the expert, and $W_{g,i}$ is the router network weight for the $i$-th expert.

\textbf{Threshold-based experts selection.}
SOTA methods~\cite{zhang2022moefication, fedus2022switch} select a fixed number of experts for all layers and inputs.
However, the necessary number of experts may differ depending on inputs and layers.
Here, we propose to select experts more adaptively for different inputs and layers for a better trade-off between model sparsity and quality, by leveraging a threshold-based method to enable adaptive selection.
The MoEfied FFN layers become:
\begin{equation}\label{eq:routing}
FFN(x) = \sum_{i=1}^n \mathbf{1}_{\{G(x)_i > \tau \}}(x) E(x)_i,
\end{equation}
where $E(x)_i$ is the output for the $i$-th expert, $\mathbf{1}(\cdot)$ is the indicator function, and $\tau$ is a predefined threshold.
Only experts with score larger than $\tau$ will be activated.
Consequently, within the same MoE layer, as the router generates larger expert scores, more experts are selected.

\subsubsection{Two-stage LTE Training}
\label{sssec:training}
Training the router poses three practical challenges. Firstly, as we consider training experts independently to mitigate bias (e.g., using a Sigmoid routing function), it is difficult to tease apart important experts from the rest. Secondly, the non-differentiability of threshold-based expert selection further complicates router training. Lastly, setting thresholds for each MoE layer is non-trivial.

To tackle these challenges, we next propose a novel two-stage training algorithm with an efficiency loss penalty to MoEfy LLMs.
Our training algorithm consists of two training stages:
\emph{a) model-router training stage} and \emph{b) model adaption stage}.

\textbf{Stage 1: Model-router training.}
In this stage, we jointly train routers and model parameters to capture the importance of experts for given inputs in the Sigmoid routers.
To address the non-differentiability issue, we switch the expert selection to a ``soft'' mode:
\begin{equation}\label{eq:soft-routing}
FFN_{soft}(x) = \sum_{i=1}^n G(x)_i E(x)_i.
\end{equation}

Instead of selecting a subset of experts, we always select all experts and multiply expert outputs $E(x)_i$ with the corresponding expert score $G(x)_i$ to make both router and model differentiable.
Since there is no discrete selection in MoE layers, all parameters are trained for each iteration.

\begin{wrapfigure}{r}{0.45\textwidth}
    \centering
    \vspace{0pt}
    \includegraphics[width=\linewidth]{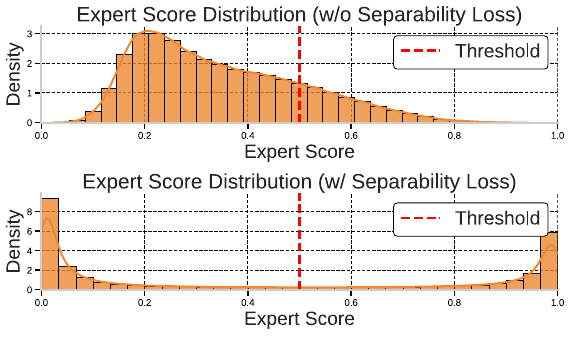}
    \caption{Expert distribution w/ and w/o separability loss. The expert scores in the model trained with the separability loss are much more separable than the model trained without the separability loss. }
    \label{fig:separability-compraison}
\end{wrapfigure}

Compared to Softmax routers, Sigmoid routers score each expert independently and do not introduce any competition on expert scores among experts.
This makes Sigmoid routers alone cannot identify more important experts in MoE layers.
As such, we design an efficiency loss penalty, $\mathcal{L}_{efficiency}$, to introduce competition among experts for Sigmoid routers:
\begin{equation}\label{eq:efficiency-loss}
\mathcal{L}_{efficiency} = \frac{1}{LN}\sum_{l=1}^L\sum_{i=1}^N |G_l(x)_i|^2,
\end{equation}
where $L$ is the number of layers, and $N$ is the number of experts in each layer.
The efficiency loss calculates the mean of expert scores across all layers.
This efficiency loss penalizes the output magnitudes of routers, driving routers to selectively allocate lower expert scores to less important experts, which helps distinguish experts with different importance.
Moreover, instead of assigning the same sparsity for all layers, the efficiency loss also induces inter-layer orchestration, allowing LTE-trained models to allocate adaptive sparsity for different layers.
We show that LTE models are capable of adaptively allocating sparsity across different layers (Figure~\ref{fig:sparsity-layer}).

For the threshold in Equation~\ref{eq:routing},
instead of choosing a threshold for each router,
we propose to select a predetermined fixed threshold and then train models to fit this threshold, i.e., train models to be threshold-aware.
Specifically, we use a threshold separability regularizer to drive models to make expert scores more separable for this given threshold:

\begin{equation}\label{eq:separability-loss}
\mathcal{L}_{separability} = \frac{1}{LN}\sum_{l=1}^L\sum_{i=1}^N \frac{1}{(G_l(x)_i - \tau)^2},
\end{equation}
where $\tau$ is a predefined threshold for all routers (we set $\tau=0.5$ for all experiments).
The separability loss encourages router outputs to diverge from the threshold $\tau$, which makes the router outputs more separable.
An example is illustrated in Figure~\ref{fig:separability-compraison}.
The only difference between the models in the two figures is if the model is trained with a separability loss.
We find that the expert scores are not separable when training without the separability loss, making it hard to decide the threshold.
However, training with the separability loss markedly improves the separability of expert scores, allowing us to choose experts with the predefined threshold.

Combined with the task performance loss $L_{task}$, our training loss in model-router training stage is:

\begin{equation}\label{eq:stage-1-loss}
\mathcal{L}_{s1} = L_{task} + \eta L_{efficiency} + \lambda L_{separability}.
\end{equation}

The efficiency coefficient, $\eta$, is used to control the trade-off between inference efficiency and task performance, and $\eta$ can be used to control the sparsity of trained models.
A higher $\eta$ indicates a larger inference cost penalty, leading to a more sparse model.
$\lambda$ is the separability loss, we set $\lambda=0.5$ for our evaluation.
We conduct an ablation study on both hyperparameters in Section~\ref{ssec:ablation}.

\textbf{Stage 2: Model Adaptation.}
To accelerate model inference, we need to switch routers to discrete selection mode (Equation~\ref{eq:routing}), and
the model trained in Stage 1 needs further training to adapt to the changes in router outputs.
In the model adaptation stage, we freeze the router parameters, switch routers to discrete selection mode (Equation~\ref{eq:routing}), and fine-tune the model to adapt to the discrete routing with the task performance loss: $\mathcal{L}_{s2} = L_{task}$.

\subsection{Hardware-efficient Implementation}
\label{ssec:hardware-implementation}

\begin{wrapfigure}{r}{0.55\textwidth}
    \centering
    \includegraphics[width=\linewidth]{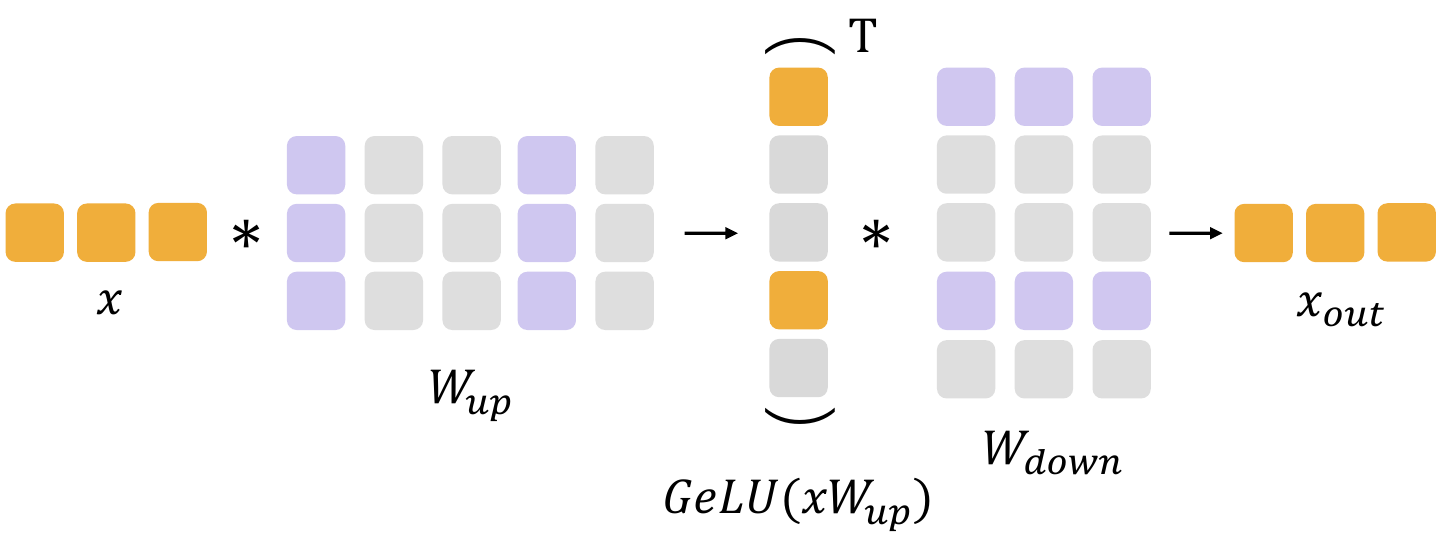}
    \caption{Structural sparsity provided by LTE in FFN layer. After selecting neurons, unselected rows and columns won't be loaded and used.
    }
    \label{fig:sparseMLP}
\end{wrapfigure}
As illustrated in Figure~\ref{fig:sparseMLP}, same as Deja Vu~\cite{liu2023deja} and moefication~\cite{zhang2022moefication}, LTE provides very structured sparsity.
After selecting neurons in intermediate layers, a CUDA kernel needs only to load the relevant columns and rows of $W_{up}$ and $W_{down}$ from GPU global memory to SRAM to compute dot product, thus reducing memory-I/O as well as computational overheads. Thanks to the structured sparsity provided by LTE, the kernel avoids non-coalesced memory access by storing column-major $W_{up}$ and row-major $W_{down}$.
In this paper, we use Triton 2.3.0 to implement a customized MLP layer to translate the sparsity to wall-clock time latency reduction.
The evaluation of our custom kernel is presented in Section~\ref{ssec:latency-reduction}.

\section{Experiment}
In this section, we conduct extensive evaluationS to verify the effectiveness of LTE.
For instance, in Section~\ref{ssec:performance-comparison},
LLaMA with LTE provides a 1.83x - 2.59x FLOPs speed-up on NLG tasks.
Along with our hardware-aware custom kernel, LTE reduces 25\% LLaMA2-7B
inference latency at 50\% sparsity.

\subsection{Experimental Setting}
\textbf{Models.}
We implement LTE on both encoder-based models~(RoBERTa$_\text{base}$, RoBERTa$_\text{large}$) and decoder-based models~(GPT2-Medium, and LLaMA-7B) with the HuggingFace's transformers.
Following previous work~\cite{zhang2022moefication}, we set each expert to contain 32 neurons in FFN layers for all models.

\textbf{Datasets:}
\textbf{1) Natural Language Understanding (NLU)}: we evaluate on eight tasks from the GLUE dataset~\cite{wang2018glue} SST-2~\cite{socher-etal-2013-recursive}, RTE~\cite{10.1007/11736790_9}, CoLA~\cite{warstadt2018neural}, MNLI~\cite{williams2018broadcoverage}, QNLI~\cite{rajpurkar2016squad}, QQP~\cite{iyer2017first}, STS-B~\cite{Cer_2017}, and MPRC~\cite{dolan-brockett-2005-automatically}.
We report the Pearson correlation coefficient for STS-B, the Matthews correlation coefficient for CoLA, and the accuracy for the other six datasets.
\textbf{2) Natural Language Generation (NLG)}:
we evaluate LTE on E2E~\cite{novikova2017e2e}, XSum~\cite{xsum-emnlp}, and Wikitext103~\cite{merity2016pointer}.
For both E2E and XSum, we report the ROUGE-L~\cite{lin2004rouge} to measure the generation text quality and report the perplexity performance for the Wikitext dataset.
\textbf{3) Instruction Tuning}:
besides downstream tasks, we also evaluate LTE on instruction tuning tasks to evaluate LTE's generalization capabilities.
We use Tulu dataset~\cite{wang2023far} to perform instruction tunning with LTE, and we evaluate models with MMLU benchmark~\cite{hendrycks2020measuring}).
MMLU benchmark is a comprehensive evaluation dataset designed to evaluate the generalization capabilities of LLMs and consists of 15908 questions from 57 distinct tasks.

\textbf{Baselines.}
We compare two baselines with LTE in our paper:
(1) \emph{MoEfication}~\cite{zhang2022moefication}: MoEfication proposes to replace soft activations with ReLU and then fine-tune the model on downstream datasets.
In our evaluation, we report the number of MoEfication with parameter clustering and ground-truth expert selection.
(2) \emph{Deja Vu}~\cite{liu2023deja}:
Deja Vu employs predictors to estimate the values of neurons in intermediate layers, pruning neurons with absolute magnitudes (absolute values for non-ReLU activations).
The models used in Deja Vu evaluation are models fine-tuned on specific tasks as well.

\vspace{-0.2cm}
\subsection{End-to-end Performance Comparison}
\label{ssec:performance-comparison}
\begin{wrapfigure}{r}{0.5\textwidth}
    \centering
    \includegraphics[width=\linewidth]{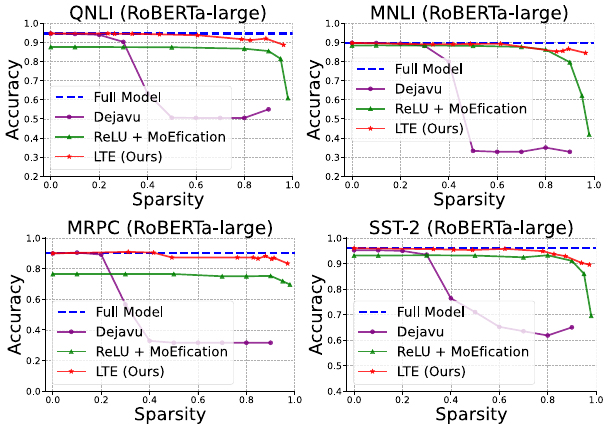}
    \caption{
    LTE consistently outperforms other baselines on 4 NLU datasets.
    }
    \label{fig:nlu-comparison}
\end{wrapfigure}
LTE, MoEfication, and Deja Vu provide the same type of structured sparsity in FFN layers.
For the convenience of comparison, we use FFN neuron sparsity as a measure of efficiency in this section.
We will further discuss how our hardware-efficient implementation translates this sparsity into wall-clock time latency reduction in Section~\ref{ssec:latency-reduction}.
For our method~(LTE), we get models with different sparsity by tuning efficiency loss $\eta$ in Equation~\ref{eq:stage-1-loss}.

\textbf{No performance drop with 80-95\% sparsity on NLU tasks.}
We first evaluate our methods on eight NLU Tasks in GLUE datasets\cite{wang2018glue} with two encoder-based models~(RoBERTa$_\text{base}$ and RoBERTa$_\text{large}$).
Due to the space limit, we report four datasets~(MRPC, SST2, QNLI, and MNLI) on RoBERTa$_\text{large}$ in Figure~\ref{fig:nlu-comparison} and other results in Figure~\ref{fig:nlu-comparison-appendix} in Appendix~\ref{app:evaluation}.

\begin{wraptable}{r}{0.5\textwidth}
\centering
\caption{GFLOPs per token for LLaMA-7B on different datasets with permitting quality drops. FLOPs of routers are also included in our method. N/A stands for the method failing to achieve the expected performance.}
\begin{tabular}{@{}lccc@{}}
\midrule
                       & XSum   & E2E    & Wiki\\ \midrule
Full                   & 12.06 & 12.06 & 12.06      \\
Deja Vu                & 7.92   & 6.42   & 7.92        \\
MoEfication            & 10.45  & N/A    & N/A          \\
R-LLaMA+MoE          & 8.27   & 7.39   & 11.1         \\
LTE (Ours)             & \textbf{5.38}   & \textbf{4.65}   & \textbf{6.59}         \\
\bottomrule
\end{tabular}
\label{tab:llama-flops}
\end{wraptable}
As shown in Figure~\ref{fig:nlu-comparison} and Figure~\ref{fig:nlu-comparison-appendix}, our method~(LTE) consistently outperforms the baselines on all datasets.
MoEfication exhibits a performance drop when sparsity reaches 70\%--80\%, and
Deja Vu experiences performance drop at a relatively low sparsity level (e.g., 30\%),
while LTE maintains good performance even at higher sparsity levels ($>$90\%).
Moreover, we notice that while ReLU-based models achieve high sparsity with a slight decrease in performance, replacing GeLU with ReLU in RoBERTa can affect performance on some datasets.
For instance, replacing GeLU with ReLU in a RoBERTa model results in around 10\% accuracy drop on the MRPC dataset.

\begin{figure}[t]
    \centering
    \includegraphics[width=\linewidth]{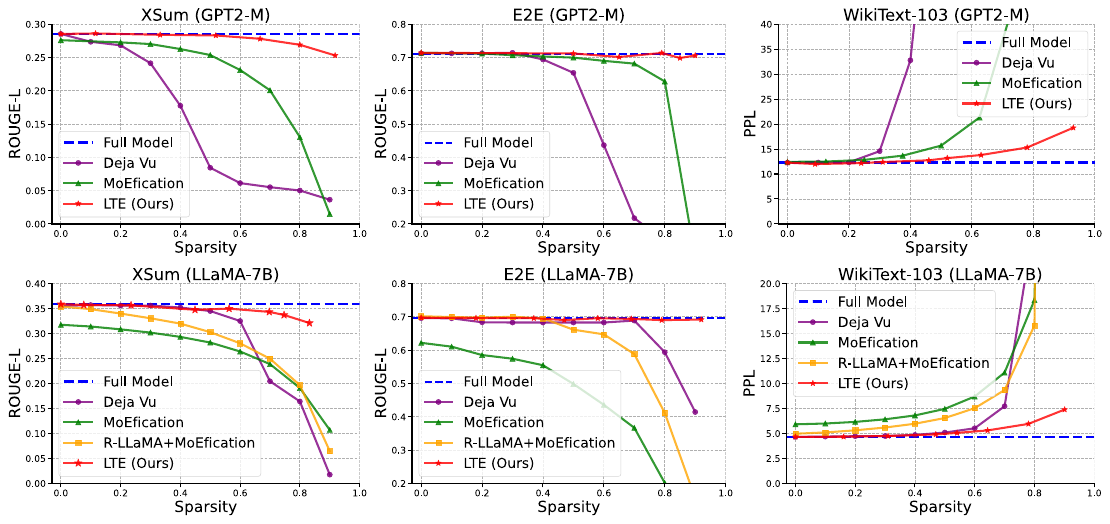}
    \vspace{-0.3cm}
    \caption{Performance comparison across three NLG datasets (each column).
    We compare LTE~(Ours) with other baselines on two decoder-based models: GPT2-M and LLaMA-7B~(each row).
    }
    \label{fig:nlg-comparison}
    \vspace{-0.3cm}
\end{figure}

\textbf{50\% FLOPs saving on NLG tasks.}
We next evaluate LTE with decoder-based models on three types of generation tasks.
Similar to NLU tasks, Figure~\ref{fig:nlg-comparison} shows that LTE outperforms all other baselines.
NLG tasks are more challenging: all methods start to have a performance drop at a lower sparsity level compared to NLU tasks.
Yet, LTE-trained models have better resilience to the sparsity increase.
Especially at a high sparsity level, while other baselines fail to generate meaningful content, LTE still generates content with a marginal quality drop.
Similarly,  we also observe the performance drop caused by replacing soft activation with ReLU.

Additionally, to better understand the computation saving of LTE, we compare the FLOPs per token of different methods on LLaMA-7B, allowing an up to 0.05 ROUGE-L decrease for XSum and E2E, and an up to 0.5 perplexity (PPL) increase for WikiText-103.
It is noteworthy that the FLOPs for routers are included in our method, constituting approximately 1\% of the total FLOPs of FFN layers.
The results are presented in Table~\ref{tab:llama-flops}.
The evaluation results show that LTE provides a 1.83x - 2.59x FLOPs speed-up and gives the most FLOPs savings among all methods.

\begin{wrapfigure}{r}{0.45\textwidth}
    \centering
    \includegraphics[width=\linewidth]{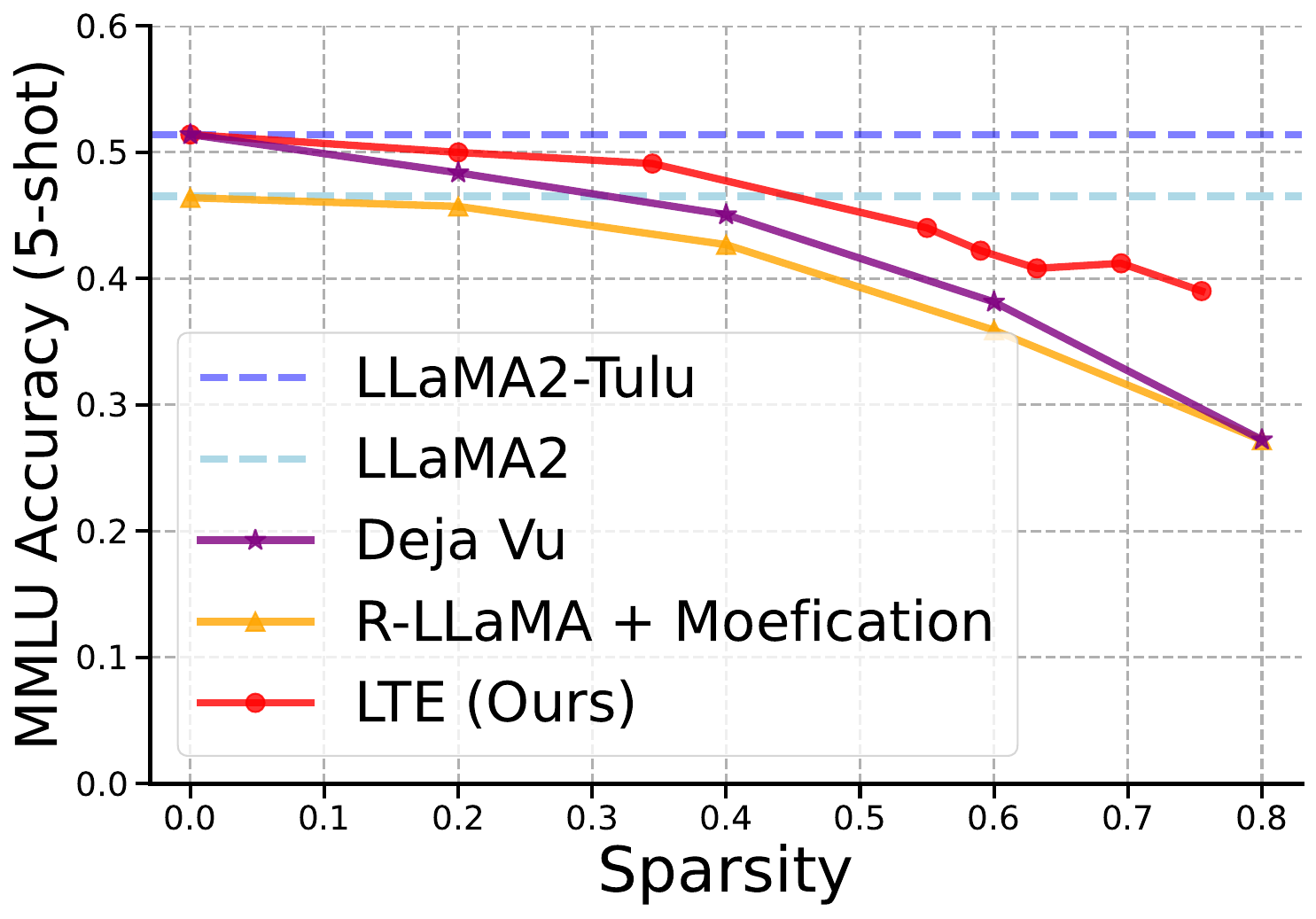}
    \caption{LTE outperforms other baselines on 5-shot MMLU benchmark.
    }
    \label{fig:mmlu-performance-comparison}
    \vspace{0.3cm}
\end{wrapfigure}
\textbf{LTE Outperforms SOTA on instruction tuning.}
To further evaluate the effectiveness of LTE, we evaluate LTE as a chatbot and verify model performance in a few-shot learning scenario.
We fine-tune LTE on the Tulu instruction tunning dataset~\cite{wang2023far}, and then evaluate the few-shot performance on MMLU benchmark~\cite{hendrycks2020measuring}.
We use LLaMA2-7B as the base model for LTE training.
For a fair comparison, we also fine-tune base models with the Tulu dataset for other baselines.
Specifically, for Deja Vu, we use Tulu-fine-tuned LLaMA2-7B as the base model.
For Moefication, we use Tulu-fine-tuned ReLULLaMA-7B as the base model to ensure the best possible baseline performance.
We present the 5-shot MMLU accuracy comparison in Figure~\ref{fig:mmlu-performance-comparison}.
The evaluation results show that LTE outperforms other baselines across all sparsity levels.
Yet, LTE shows a weaker sparsity-performance trade-off compared to downstream tasks.
Our hypothesis is that LTE routers change the structure of the original LLaMA model, which influences general language patterns learned by models in pretraining.
Instruction tuning alone is not enough to fully recover this ability (considering instruction tuning has much smaller datasets).
We believe that training with a wider variety of data (like pretraining data) will further improve LTE performance.

\subsection{Translate Sparsity to Wall-clock Time Speedup}
\label{ssec:latency-reduction}

\begin{wrapfigure}{r}{0.63\textwidth}
    \centering
    \includegraphics[width=\linewidth]{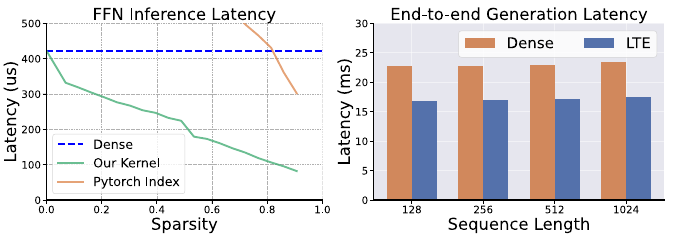}
    \caption{
    Wall-clock time latency comparison on dense FFN blocks and our custom Triton kernel~(Left).
    End-to-end generation latency comparison between dense model and LTE with 50\% sparsity~(Right).
    }
    \label{fig:latency-comparison}
\end{wrapfigure}
\textbf{25\% wall-clock time latency reduction.}
As discussed in Section~\ref{ssec:hardware-implementation}, the sparsity introduced by LTE is highly structured and can be easily translated to latency reduction.
This section demonstrates that our custom Triton kernel effectively translates the LTE sparsity to wall-clock time speed up.
We evaluate wall-clock time speed up using LLaMA2-7B on a single 3090Ti.
We also reported vanilla Pytorch indexed matrix multiplication.
Due to the requirement for new memory allocations, slicing in Pytorch is very time-consuming. This makes indexed matrix multiplication even slower than dense matrix multiplication at most sparsity levels.
We report the wall-clock time latency comparison in Figure~\ref{fig:latency-comparison}~(noting that the router inference latency is already included in LTE latency).
Compared to a dense FFN block, our custom Triton kernel achieves nearly linear speed-up with respect to sparsity (Left).
At around 50\% sparsity, LTE reduces end-to-end generation latency by approximately 25\%.

\subsection{Addtional Comparison}

\textbf{Model pruning.}
As discussed in Section~\ref{sec:lte-related-work}, besides dynamic contextual sparsity provided by LTE, static sparsity provided by model pruning is also a common method to accelerate model inference.
In this section, we present a performance comparison between Wanda~\cite{sun2023simple} and LTE in Table~\ref{tab:model-pruning-compare}.
The sparsity reported here is the overall sparsity of the entire model, which is different from the FFN sparsity reported in Seciton~\ref{ssec:performance-comparison}.
For LTE, we recalculate the overall sparsity based on the FFN sparsity.
The evaluation results show that LTE achieves a lower perplexity than Wanda.
The evaluation results show that, given a similar sparsity, LTE achieves better perplexity.

\begin{table}[t]
    \centering
    \footnotesize
    \caption{Performance comparison with model pruning. We apply Wanda~\cite{sun2023simple} on WikiText103 fine-tuned LLaMA-7B with the author-implemented code. Besides using C4 as the calibration data suggested by the Wanda paper, we also try to use WikiText for calibration to improve this baseline. The sparsity reported here is the overall sparsity of the entire model, which is different from the FFN sparsity. The evaluation results show that LTE achieves a lower perplexity than Wanda.}
    \label{tab:model-pruning-compare}
    \begin{tabular}{lccccccc}
        \toprule
        Method & \begin{tabular}[c]{@{}c@{}}Wanda \\ (2:4)\end{tabular} & \begin{tabular}[c]{@{}c@{}}Wanda \\ (4:8)\end{tabular} & \begin{tabular}[c]{@{}c@{}}Wanda \\ (Unstructured)\end{tabular} & \begin{tabular}[c]{@{}c@{}}Wanda \\ (2:4)\end{tabular} & \begin{tabular}[c]{@{}c@{}}Wanda \\ (4:8)\end{tabular} & \begin{tabular}[c]{@{}c@{}}Wanda \\ (Unstructured)\end{tabular} & \begin{tabular}[c]{@{}c@{}}LTE \\ ($\eta=2$)\end{tabular}\\
        \midrule
        Overall Sparsity & 50\% & 50\% & 50\% & 50\% & 50\% & 50\% & 52\% \\
        Calibration Data & C4 & C4 & C4 & Wiki & Wiki & Wiki & - \\
        PPL & 11.45 & 8.39 & 7.04 & 10.82 & 8.17 & 6.87 & \textbf{5.95} \\
        \bottomrule
    \end{tabular}
\end{table}

\subsection{Ablation Study}
\label{ssec:ablation}

\begin{figure}[htbp]
    \centering
    \begin{minipage}[t]{0.32\textwidth}
        \centering
        \includegraphics[width=\linewidth]{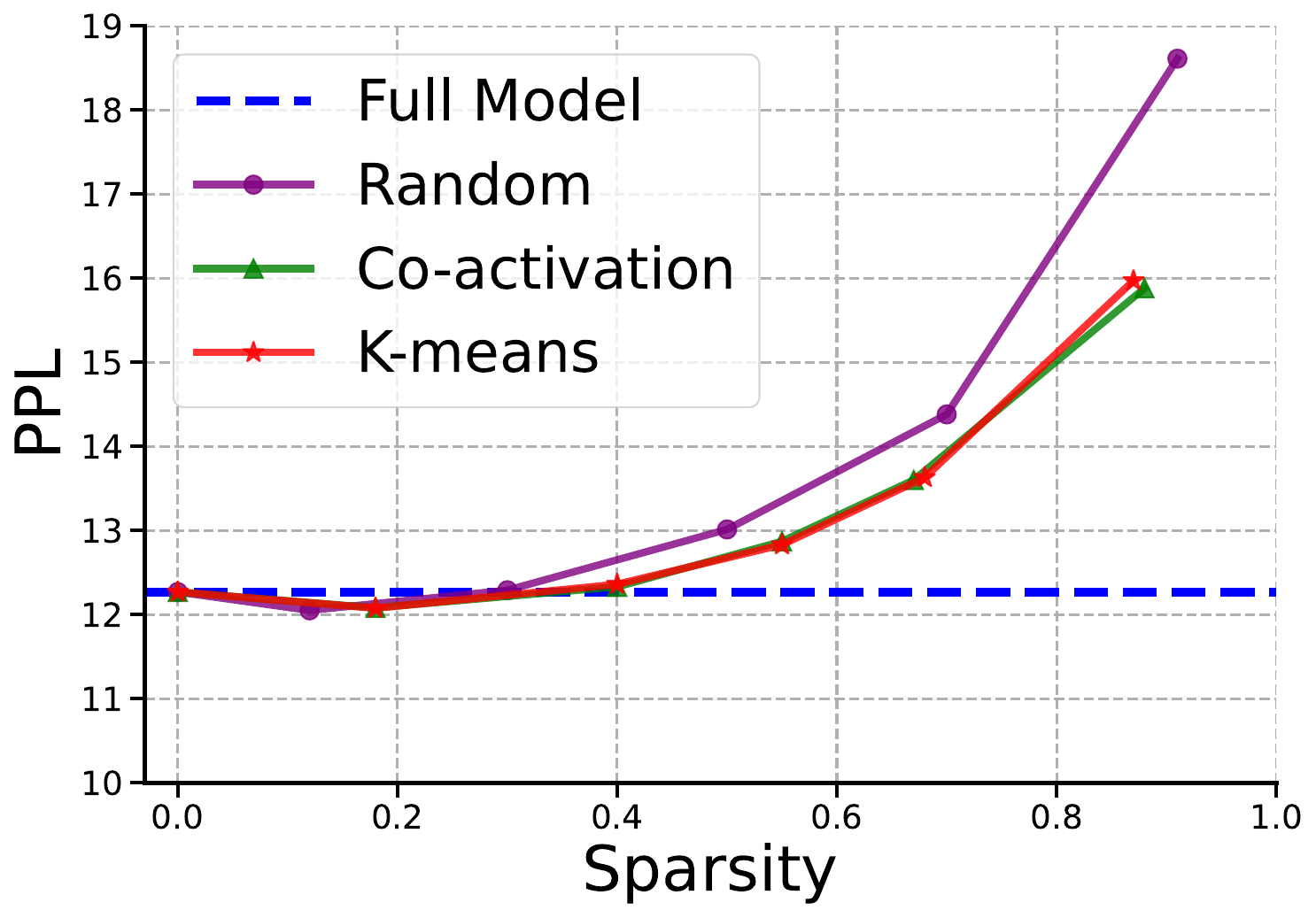}
        \caption{Performance comparison between different experts grouping algorithms.}
        \label{fig:ablation-experts-grouping}
    \end{minipage}
    \hfill %
    \begin{minipage}[t]{0.32\textwidth}
        \centering
        \includegraphics[width=\linewidth]{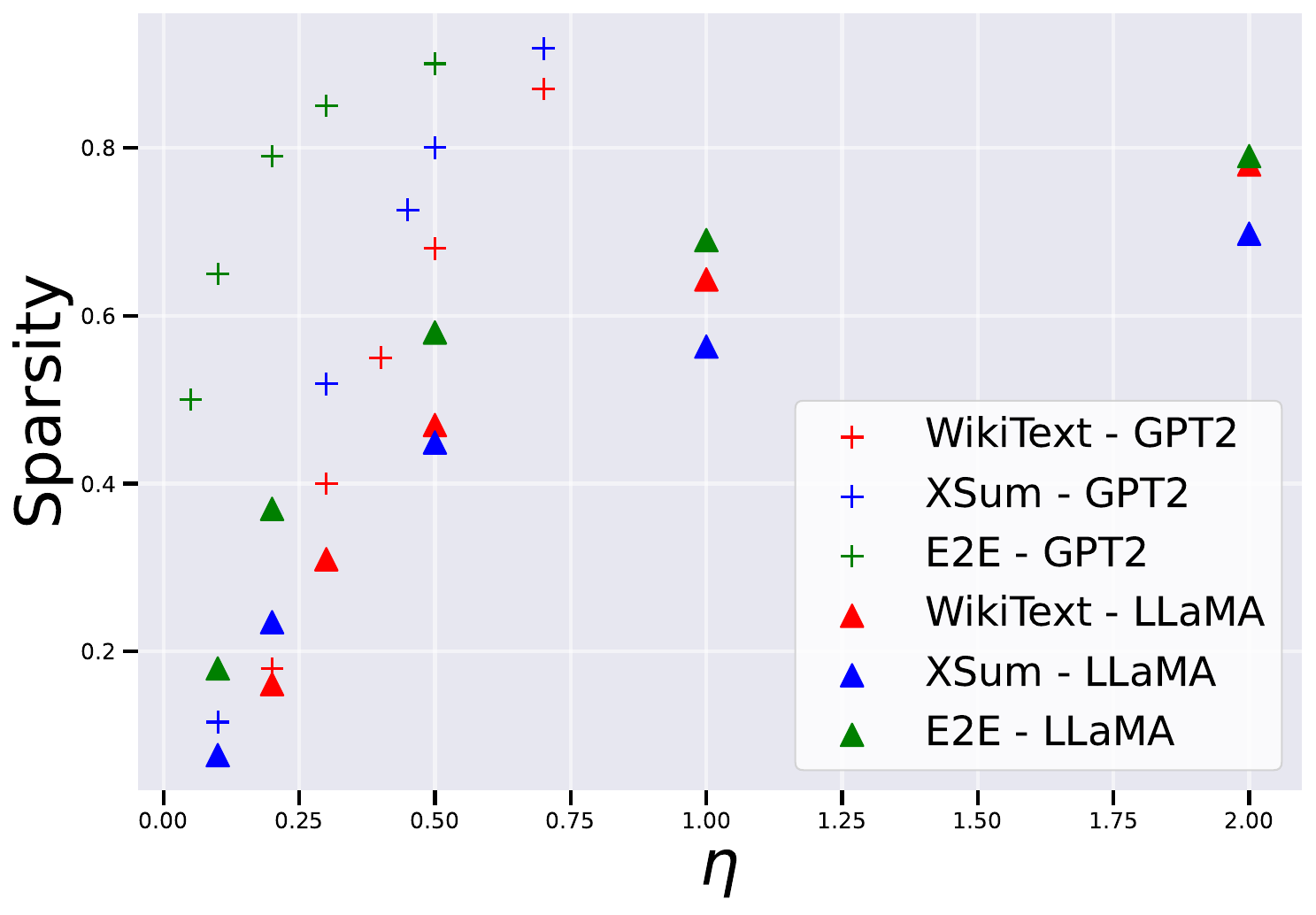}
        \caption{The relation between $\eta$ and sparsity on different models and tasks.}
        \label{fig:ablation-eta}
    \end{minipage}
    \hfill %
    \begin{minipage}[t]{0.32\textwidth}
        \centering
        \includegraphics[width=\linewidth]{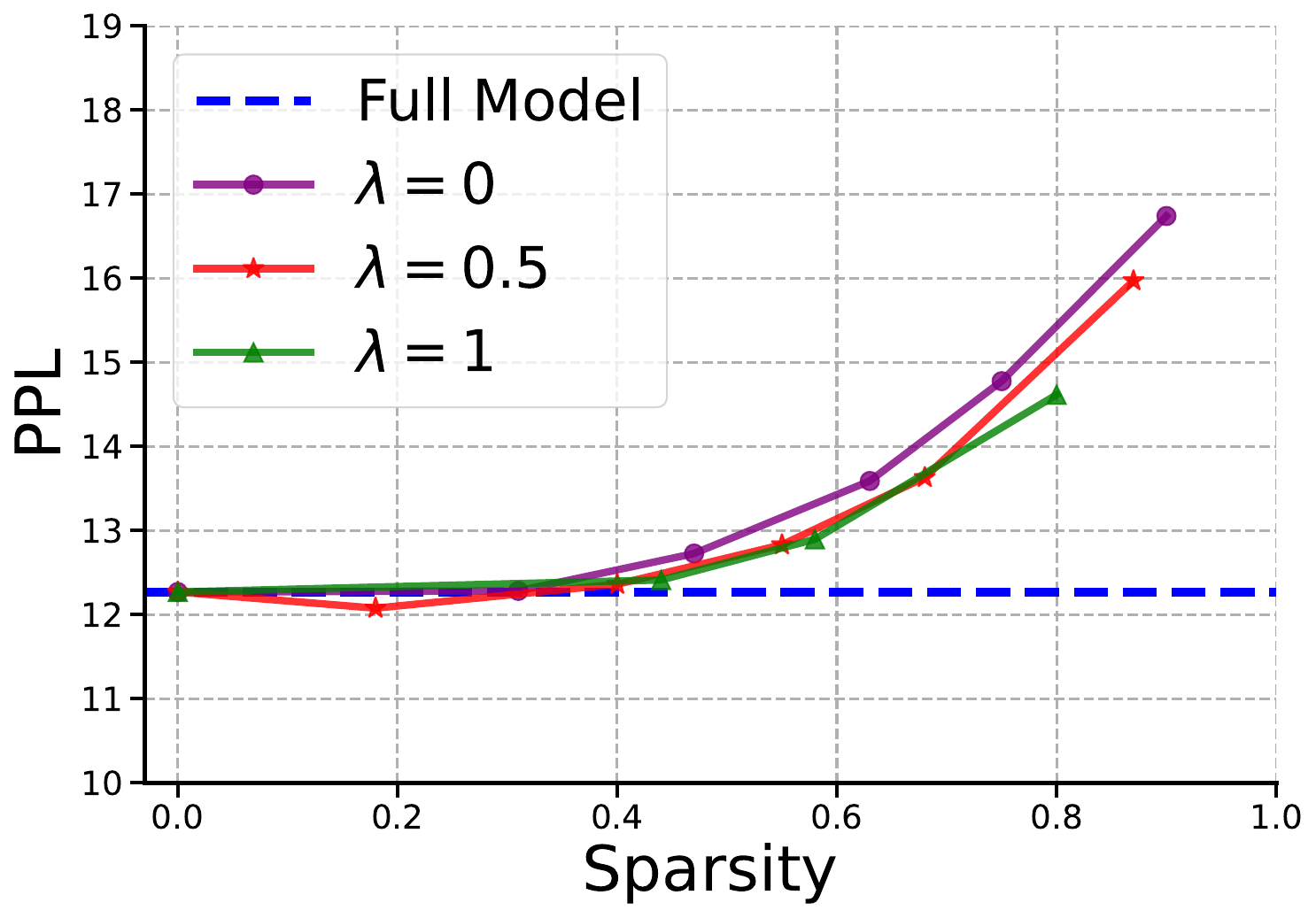}
        \caption{Performance comparison between different $\lambda$ on WikiText-103~(GPT2-M).}
        \label{fig:ablation-sep}
    \end{minipage}
    \vspace{0.3cm}
\end{figure}

\textbf{Experts grouping algorithms.}
Three different expert grouping algorithms are explored in \cite{zhang2022moefication}: random grouping, parameter K-means clustering, and co-activation graph split. The co-activation graph split algorithm constructs a graph where each neuron represents a node and the edges denote the frequency of co-activation between neuron pairs. This graph is subsequently divided into subgraphs using the graph split algorithm \cite{karypis1998fast}, with each subgraph representing a group of neurons. To study the effectiveness of these grouping algorithms, we compare their LTE performance by fine-tuning GPT2 on the WikiText-103 dataset.
As shown in Figure~\ref{fig:ablation-experts-grouping}, both co-activation graph split and parameter K-means grouping have very similar performance~(similar to finding in \cite{zhang2022moefication}) and outperform random grouping. Given that parameter K-means grouping is simpler to implement, as co-activation graph split requires collecting activation data, we use K-means grouping in our paper.

\textbf{The relation between efficiency loss hyperparameter $\eta$ and sparsity.}
As discussed in Section~\ref{sssec:training}, efficiency loss hyperparameter $\eta$ is used to balance the trade-off between task performance and sparsity~(efficiency).
We illustrate the relation between $\eta$ and sparsity across different models and tasks in Figure~\ref{fig:ablation-eta}.
For a given task and model, increasing $\eta$ results in greater sparsity.

\begin{wrapfigure}{r}{0.6\textwidth}
    \centering
    \includegraphics[width=\linewidth]{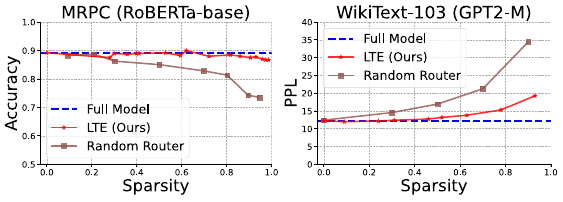}
    \caption{Performance comparison between random routers and routers trained with LTE. Models with LTE-trained routers consistently outperform models with random routers.
    }
    \label{fig:ablation-random-router}
    \vspace{0.2cm}
\end{wrapfigure}
\textbf{Ablation study on separability loss.}
Another hyperparameter in LTE training, $\lambda$, is used to encourage the separability of router outputs, which facilitates us to choose the threshold to pick neurons (as shown in Figure~\ref{fig:separability-compraison}).
We compare models trained with different lambda in Figure~\ref{fig:ablation-sep}.
When increasing $\lambda$ from $0$ (i.e., no separability loss) to $0.5$, we observe a constant perplexity drop for all sparsity.
However, if we keep increasing $\lambda$, perplexity does not further decrease.

\begin{wrapfigure}{r}{0.4\textwidth}
    \centering
    \includegraphics[width=\linewidth]{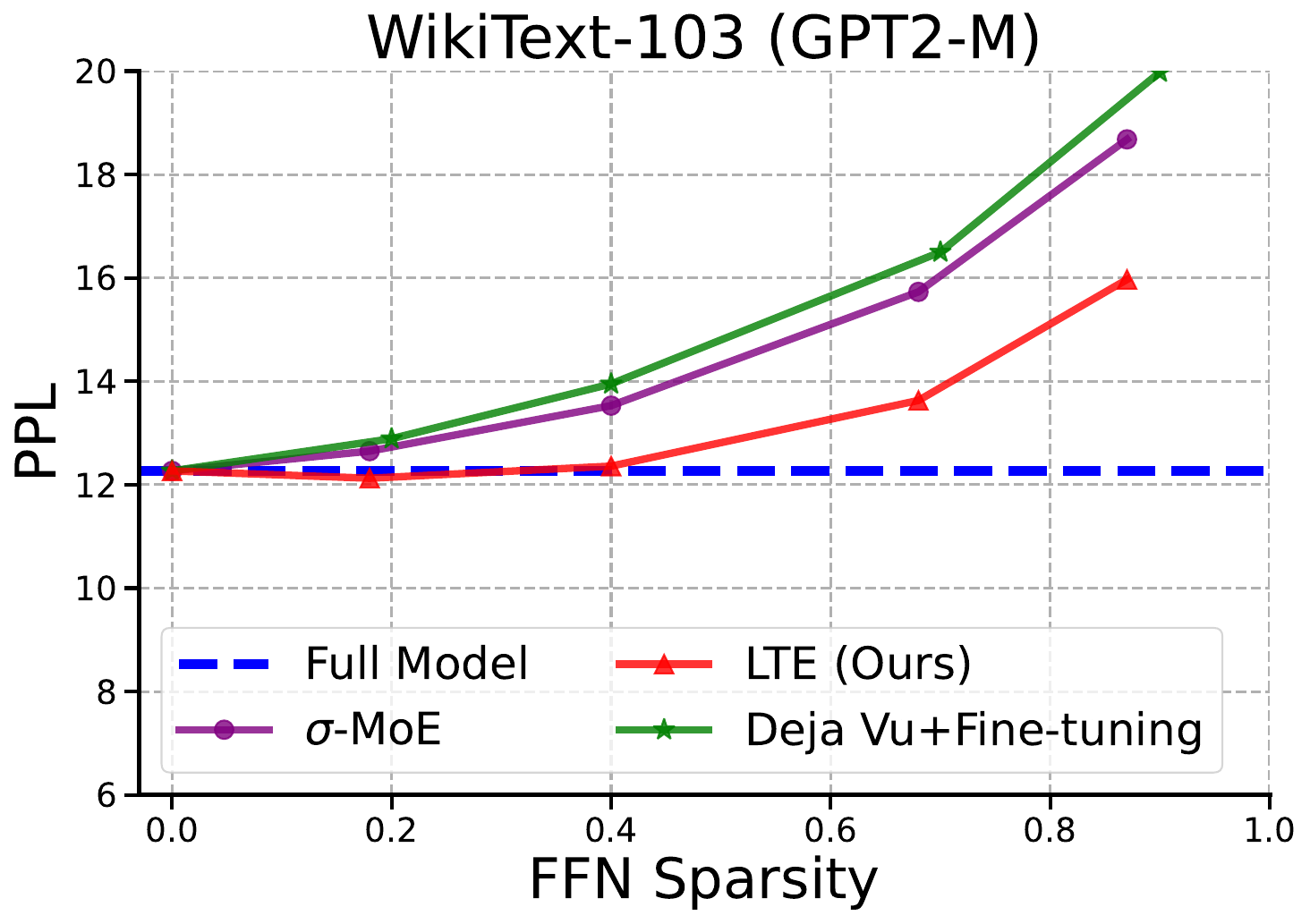}
    \caption{Performance comparison with new baselines: 1) $\sigma$-MoE~\cite{csordas2023approximating} and 2) Deja VU~\cite{liu2023deja} with further fine-tuning. We evaluate two baselines by fine-tuning GPT2 medium with WikiText103.
    LTE outperforms the other two baselines across different sparsity.
    }
    \label{fig:sigmoid-moe-comparison}
\end{wrapfigure}
\textbf{Effectiveness of LTE-trained routers.}
We next demonstrate the effectiveness of LTE's first training stage, the model-router training stage, in helping LLMs gain better sparsity.
To build a baseline, instead of using routers trained in the model-router training stage, we use randomly initialized networks as routers and train LLMs to adapt to random routers.
Since we cannot control the sparsity of LLMs with a threshold in a random router, we decide the sparsity for MoE layers by picking K top experts: we first fix a K to decide the sparsity and then start to train the model.
Figure~\ref{fig:ablation-random-router} compares the performance of models with different routers.
We observe that routers trained with LTE significantly contribute to achieving a better trade-off between sparsity and model performance, which proves the effectiveness of the model-router training stage.

\textbf{Other baselines with further fine-tuning.}
A key distinction between LTE and other baselines is that LTE requires additional fine-tuning to make the model efficiency-aware. In this section, we assess whether further fine-tuning can enhance the performance of other baselines, aiming to understand the contribution of additional training to LTE's effectiveness. Specifically, we fine-tune two baselines on GPT2-Medium using the WikiText103 dataset:
1) we first fine-tune GPT2-M with Wikitext and apply Deja Vu on the fine-tuned models.
Then, we further fine-tune MoEfied models with WikiText103.
2) We implement $\sigma$-MoE~\cite{csordas2023approximating} on GPT2-Medium model and fine-tune models with the WikiText103 dataset.
For both baselines, we train models for the same training time as we train LTE models.
The evaluation results shown in Figure~\ref{fig:sigmoid-moe-comparison} show that LTE outperforms the other two baselines across different sparsity levels.

\section{Conclusion}
In this paper, we explore how to enable better activation sparsity for fast LLM inference.
We introduce a novel algorithm, LTE, for the model to automatically learn to activate fewer neurons, by grouping neurons into different experts and then adaptively selecting important ones for specific inputs and layers.
Our extensive evaluations, spanning four LLMs and eleven datasets, show that LTE can achieve 1.83x - 2.59x FLOPs speed-up on LLaMA, thus execution efficiency, outperforming state-of-the-art designs.
We believe that our work can inspire more research on designing more efficiency-aware training methods, making LLMs more accessible to the broad community.

\section*{Acknowledgement}
My work was partially supported by DARPA Grant HR00112020008, National Science Foundation Grant No. 2039445, and Cisco Research Grant. 
Any opinions, findings, and conclusions or recommendations expressed in this material are those of the author(s) and do not necessarily reflect the views of our research sponsors.

\bibliographystyle{plain}

\newpage
\appendix
\section*{Broader Impact}

This paper presents the Learn-To-be-Efficient (LTE) algorithm to accelerate the LLM inference.
A more efficient LLM inference can significantly enhance the accessibility and sustainability of LLMs.
By reducing computational demands, it addresses environmental concerns through lower energy consumption and helps democratize access to advanced AI technologies. 
We believe that our work can better help smaller organizations, educational institutions, and researchers, who previously faced barriers due to resource limitations, access LLMs more easily.
To sum up, the LTE algorithm not only advances the field technically but also broadens the scope of AI's benefits to a wider, more inclusive community.

\section*{Limitations}
While LTE outperforms other baselines, LTE needs further fine-tuning to get contextual sparse models. 
(Yet, training is a one-time investment that yields long-term benefits for LLM serving.)
A future study is to combine LTE with other parameter-efficient fine-tuning techniques to reduce LTE training overhead.
Another limitation is our computational resource, which prevents us from training LTE on pretraining or RLHF.
As discussed in Section~\ref{ssec:performance-comparison}, we believe that training with a wider variety of data will further improve LTE performance. We leave this for future exploration.

\section{Code and Datasets License}
\label{sec:appendix_code}
\textbf{Codebase}.
Our model implementation is based on Huggingface transformer repo: 
\href{https://github.com/huggingface/transformers/blob/main/LICENSE}{Apache-2.0 license}

\textbf{Datasets}. We list the license of used datasets as follow:

GLUE dataset~\cite{wang2018glue}: 
\href{https://gluebenchmark.com/faq}{Custom License};

WikiText103~\cite{merity2016pointer}: 
\href{https://paperswithcode.com/dataset/wikitext-103}{CC BY-SA 3.0};

XSum~\cite{xsum-emnlp}: 
\href{https://github.com/EdinburghNLP/XSum/blob/master/LICENSE}{MIT License};

E2E~\cite{novikova2017e2e}: 
\href{https://creativecommons.org/licenses/by-sa/4.0/}{CC4.0-BY-SA};

Tulu~\cite{wang2023far}: 
\href{https://opendatacommons.org/licenses/by/1-0/}{ODC-BY};

MMLU~\cite{hendrycks2020measuring}: 
\href{https://github.com/hendrycks/test#citation}{Custom License}.

\section{Experiment Settings}
We presented all training hyperparameters in Table~\ref{tab:hyper-roberta-base},~\ref{tab:hyper-roberta-large},~\ref{tab:hyper-gpt2}, and~\ref{tab:hyper-llama}.
For hyperparameters presented in \{\}, we select the best hyperparameter for each task.
We follow the settings in RoBERTa paper~\cite{liu2019roberta} to fine-tune RoBERTa on GLUE datasets. 
We set the coefficient for the separability loss ($\lambda$ in Equation~\ref{eq:stage-1-loss}) to be 0.5 for all stage 1 training.
We use different $\eta$ to control the sparsity of trained models (Figure~\ref{fig:ablation-eta}). 
Hardware: All models are trained and evaluated on A100, A40, and 3090Ti, depending on memory usage and availability.

\section{Additional Evaluation Results}
\label{app:evaluation}

\subsection{NLU Performance Comparison}
\begin{figure*}[h]
    \centering
    \includegraphics[width=\linewidth]{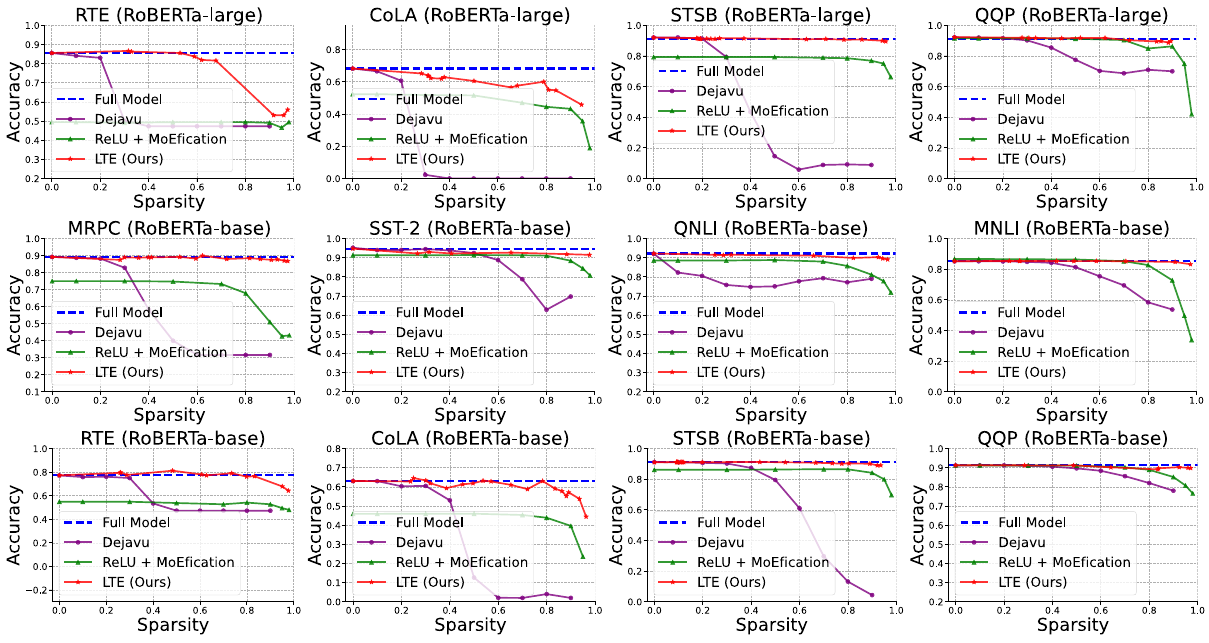}    
    \caption{Performance comparison across four NLU datasets from GLUE dataset. 
    The evaluation results show that LTE consistently outperforms other baselines in each dataset for both models.}
    \label{fig:nlu-comparison-appendix}
    \vspace{-0.4cm}
\end{figure*}

In Figure~\ref{fig:nlu-comparison-appendix}, we present the evaluation results of the rest four NLU datasets. 
We have similar findings as we discussed in Section~\ref{ssec:performance-comparison}: LTE achieves a better trade-off between sparsity and task performance.
However, we notice that, LTE outperforms MoEfication on the CoLA dataset, but has worse performance than KLA at low sparsity level. 
The size of CoLA is relatively small and may not be efficient for models to learn good routers.

\subsection{Further Analysis on Sparsity Provided by LTE}

\begin{wrapfigure}{r}{0.35\textwidth}
    \centering
    \includegraphics[width=\linewidth]{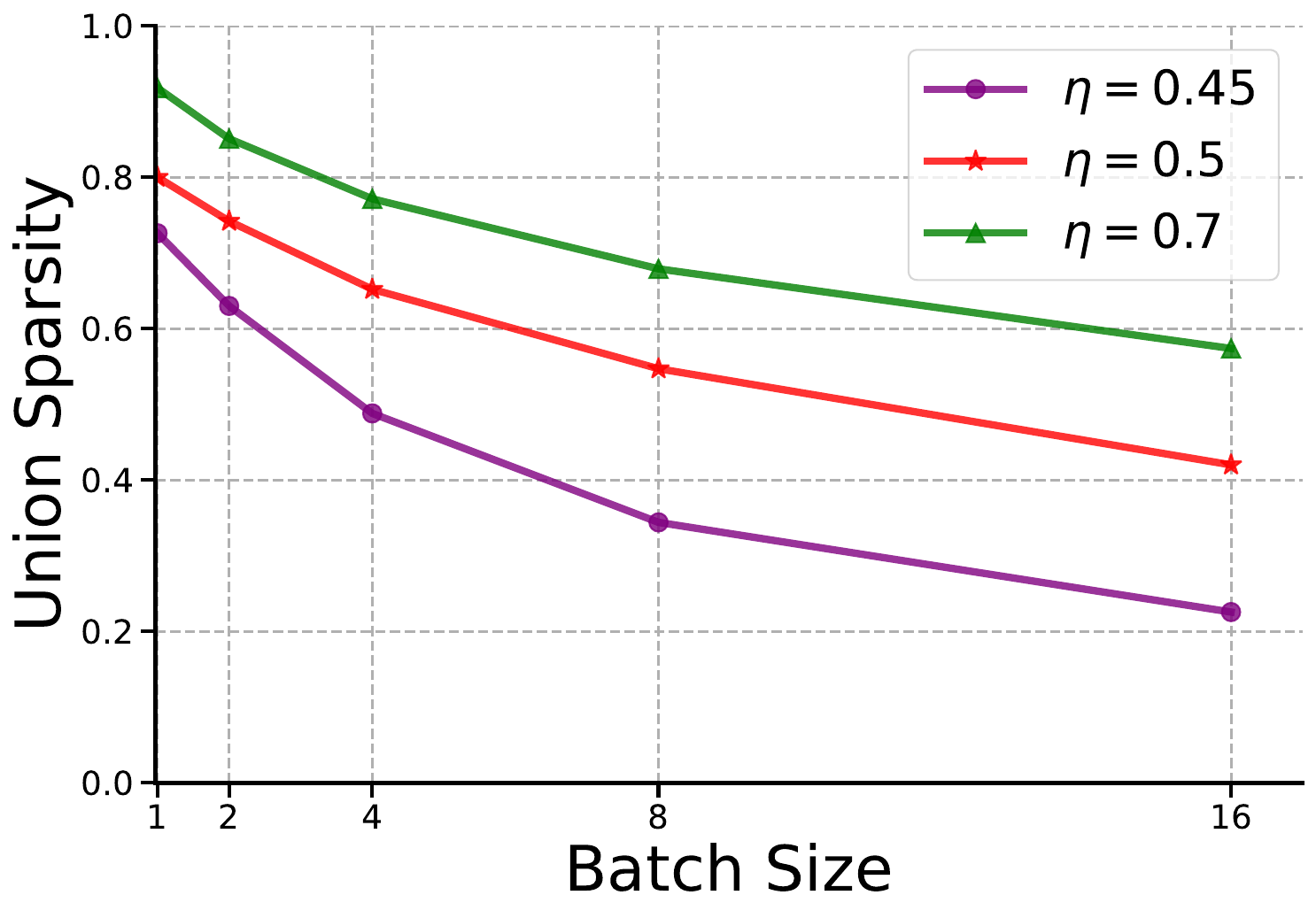}
    \caption{Union sparsity changes w.r.t. batch size. 
    }
    \label{fig:analysis-union-sparsity}
\end{wrapfigure}
\textbf{Contextual Sparsity for Larger Batches.}
Similar to previous contextual sparsity work~\cite{liu2023deja}, we observe that as the batch size increases, more neurons are activated, leading to a reduction in union sparsity.
These dynamics are illustrated in 
Figure\ref{fig:analysis-union-sparsity}.
Similar to Deja Vu~\cite{liu2023deja}, 
the number of activated neurons does not grow linearly with the batch size, indicating a non-uniform distribution of parameter access from different inputs.
This property can extend LTE for a high-throughput batch setting by batching inputs activating similar neurons together to achieve a high union sparsity.

\begin{wrapfigure}{r}{0.5\textwidth}
    \centering
    \includegraphics[width=\linewidth]{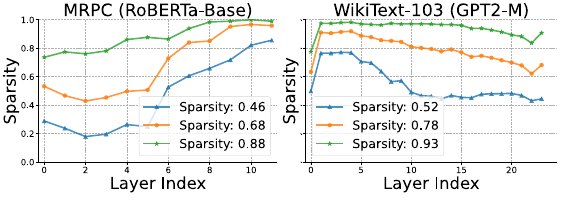}
    \vspace{-.1cm}
    \caption{
    Sparsity across different layers. 
    We present the sparsity of different layers in an encoder-based model~(RoBERTa$_\text{base}$) for MRPC and a decoder-based model~(GPT2-M) for WikiText103. 
    Different layers learn to have different sparsity, and RoBERTa$_\text{base}$ and GPT2-M have different sparsity patterns across layers.
    }
    \label{fig:sparsity-layer}
\end{wrapfigure}

\textbf{Adaptive Sparsity across Layers.}
Instead of assigning a fixed sparsity to each layer, LTE uses the efficiency loss to introduce competition across different layers to allocate the computation budget to more important layers.
Here, we conduct a study on the distribution of sparsity across layers in models trained with LTE.
As shown in Figure~\ref{fig:sparsity-layer}, we observe that LTE achieves different sparsity levels across different layers.
For example, the MRPC~(RoBERTa$_\text{base}$) model with 0.46 sparsity has 0.2 sparsity at layer 2, but has more than 0.8 sparsity at layers 10 and 11.

Another intriguing finding is that RoBERTa$_\text{base}$ and GPT2-M have very different sparsity patterns.
In RoBERTa, the sparsity of MoE layers significantly increases with depth, contrasting with GPT2-M, where MoE layer sparsity decreases as layer depth increases. 
This difference can stem from differences in the design of these two transformers and tasks.
Given that MRPC is a sentence classification task, encoder-based transformers like RoBERTa primarily rely on the "[CLS]" token for classification. 
Consequently, other tokens become less important in deeper layers, allowing sparser FFN layers.
In contrast, GPT2, a decoder-based transformer designed for language generation, requires all tokens to generate the next token. 
Furthermore, recent research~\cite{geva2021transformer} on interpretability suggests that deeper FFN layers store more complex semantic information, which may drive deep FFN layers to be less sparse.

\clearpage

\begin{table}
\centering
\caption{
RoBERTa-base hyperparameter for LTE training on GLUE dataset.
}
\begin{tabular}{lccc}
\hline
\textbf{Hyperparameter}  & \textbf{LTE-Stage 1} & \textbf{LTE-Stage 2}\\
\hline
Learning rate &  \{2e-5, 5e-5\} & \{5e-5\}\\
Training batch size &  \{16, 32\}  & \{16, 32\}\\
Training epochs &  \{1, 10\} & \{1, 3, 10\}\\
Weight decay & 0 & 0\\
Warm up ratio &  0.06 & 0.06\\
\hline
\end{tabular}
\label{tab:hyper-roberta-base}
\end{table}

\begin{table}
\centering
\caption{
RoBERTa-large hyperparameter for LTE training on GLUE dataset.
}
\begin{tabular}{lccc}
\hline
\textbf{Hyperparameter}  & \textbf{LTE-Stage 1} & \textbf{LTE-Stage 2}\\
\hline
Learning rate &  \{2e-6, 5e-6, 3e-5\} & \{2e-6, 5e-6, 3e-5\}\\
Training batch size & \{16, 32\}  & \{16, 32\}\\
Training epochs &  \{1, 2, 3, 10\} & \{3, 5, 10\}\\
Weight decay &  0 & 0\\
Warm up ratio & 0.06 & 0.06\\
\hline
\end{tabular}
\label{tab:hyper-roberta-large}
\end{table}

\begin{table}
\centering
\caption{
GPT2-Medium hyperparameter for fine-tuning and LTE training on XSum, E2E, and WikiText-103 datasets. (Numbers in the bracket correspond to three datasets in order.)
}
\begin{tabular}{lccc}
\hline
\textbf{Hyperparameter} & \textbf{Fine-tuning}  & \textbf{LTE-Stage 1} & \textbf{LTE-Stage 2}\\
\hline
Learning rate & 5e-5 & 5e-5 & 5e-5 \\
Training batch size & 8 & 8  & 8\\
Training epochs & 3 & 1 & (1, 1, 3)\\
Weight decay & 0 & 0 & 0\\
Warm up ratio & 0.06 & 0.06 & 0.06\\
\hline
\end{tabular}
\label{tab:hyper-gpt2}
\end{table}

\begin{table}
\centering
\caption{
LLaMA hyperparameter for fine-tuning and LTE training on XSum, E2E, and WikiText-103 datasets.
}
\begin{tabular}{lccc}
\hline
\textbf{Hyperparameter} & \textbf{Fine-tuning}  & \textbf{LTE-Stage 1} & \textbf{LTE-Stage 2}\\
\hline
Learning rate & 1.5e-5 & 1.5e-5 & 1.5e-5 \\
Training batch size & 16 & 16  & 16\\
Training epochs & 3 & 1 & 1 \\
Weight decay & 0 & 0 & 0\\
Warm up ratio & 0.06 & 0.06 & 0.06\\
\hline
\end{tabular}
\label{tab:hyper-llama}
\end{table}

\begin{table}
\centering
\caption{
Hyperparameters for instruction tunning with Tulu dataset.
}
\begin{tabular}{lccc}
\hline
\textbf{Hyperparameter} & \textbf{Fine-tuning}  & \textbf{LTE-Stage 1} & \textbf{LTE-Stage 2}\\
\hline
Learning rate & 2e-5 & 2e-5 & 2e-5 \\
Training batch size & 128 & 128  & 128\\
Training epochs & 3 & 1 & 4 \\
Weight decay & 0 & 0 & 0\\
Warm up ratio & 0.06 & 0.06 & 0.06\\
\hline
\end{tabular}
\label{tab:hyper-tulu}
\end{table}

\clearpage

\end{document}